\documentclass[10pt,letterpaper]{article}

\usepackage[margin=1.0in,top=0.9in,bottom=1.0in,headheight=12pt,headsep=15pt]{geometry}
\usepackage[T1]{fontenc}
\usepackage[utf8]{inputenc}
\usepackage{lmodern}
\usepackage{amsmath,amssymb}
\usepackage{graphicx}
\usepackage{subcaption}
\usepackage{natbib}

\usepackage{oxflock}          % defines the palette used by hyperref below

\usepackage[
  colorlinks=true,
  linkcolor=oxfordmid, citecolor=oxfordmid, urlcolor=oxfordmid,
  pdftitle={Euston: Training Away Mathematical Sycophancy Without Losing the Mathematics},
  pdfauthor={Zehua Cheng, Wei Dai, Jiahao Sun},
  pdfkeywords={GRPO, reinforcement learning, sycophancy, claim verification, GraphSynth}
]{hyperref}
\usepackage{url}
\selectfont
\newcommand{\pp}{\,pp}
\newcommand{\ci}[2]{\text{95\% CI }[#1,\,#2]}
\newcommand{\best}[1]{\textbf{#1}}

\reportkicker{Technical Report}
\reporttitle{Euston: Training Away Mathematical Sycophancy\\ Without Losing the Mathematics}
\runningtitle{Euston: mathematical sycophancy under GRPO}
\runningauthor{Cheng, Dai \& Sun}
\reportauthor{Zehua Cheng}{1}
\reportauthor{Wei Dai}{2}
\reportauthor{Jiahao Sun}{2}
\reportaffil{1}{University of Oxford}
\reportaffil{2}{FLock.io}
\reportcorresp{ai@flock.io}
\reportdate{19 September 2026}
\reportkeywords{Euston; reinforcement learning from verifiable rewards; GRPO; sycophancy;
mathematical claim verification; synthetic data generation; capability retention}

\begin{document}

\makereporttitle{%
Reasoning language models are trained to produce solutions, not to refuse them, and this
bias persists when the problem they are handed is false. Asked to prove a corrupted theorem,
a strong model will typically comply and produce a confident derivation of something untrue.
We present Euston, an 8B mathematical claim-verification model trained to resist exactly this. Training data
were generated with GraphSynth, a probabilistic factor-graph generator that couples attribute-level
diversity to decode-time structural masking and span-synchronized verification, yielding 3{,}026
matched true/corrupted statement pairs (6{,}052 statements) drawn from arXiv papers spanning
2010--2025. We fine-tuned DeepSeek-R1-0528-Qwen3-8B with GRPO under a rule-based,
zero-API reward for 189 steps on four H100 GPUs. On a balanced 200-true/200-false held-out split,
balanced accuracy rises from 29.50\% to 63.75\% and the discrimination gap---the difference between
the rate of calling false statements false and the rate of calling true statements false---moves from
$-0.5$\pp{} ($z=-0.1$) to $+27.5$\pp{} ($z=+6.0$). Critically, the gain is not purchased with general
mathematical ability: AIME 2026 accuracy under official semantics is 65.00\% against a 69.17\% base,
a difference of $-4.17$\pp{} that is not statistically significant, whereas an earlier run of the same
recipe on a smaller GraphSynth corpus collapsed to 40.00\%. Median response length also falls from 19{,}217 to 18{,}296
tokens and the truncation rate from 25.8\% to 8.3\%, so the improvement does not come from thinking
longer. We report the result together with the confounds that bound its interpretation, principally
the all-false composition of the official evaluation sets and the low precision implied at realistic
error prevalence.
}

%%%%%%%%%%%%%%%%%%%%%%%%%%%%%%%%%%%%%%%%%%%%%%%%%%%%%%%%%%%%%%%%%%%%%%%%%%%%%%%
\section{Introduction}

A reasoning model that has been optimised to solve problems learns, among other things, that
producing an answer is rewarded and that producing nothing is not. When the problem itself is
sound this is exactly the behaviour one wants. When the problem is unsound---a lemma whose
hypothesis has been quietly weakened, a bound with a flipped inequality, a constant altered by a
factor of two---the same disposition becomes a liability, because the model will supply a fluent
derivation of a false statement rather than report that the statement cannot be derived. This is
the mathematical instance of the sycophancy failure documented more broadly in aligned language
models \citep{sharma2024sycophancy}, and it matters in proportion to how much we intend to use such
models as verifiers rather than as solvers.

The BrokenArXiv family of benchmarks makes the failure measurable by pairing statements taken from
real arXiv papers with deliberately corrupted variants and asking the model to adjudicate
\citep{brokenarxiv}. Two things make progress on this benchmark harder than it first appears. The
first is that the naive fix---reward the model for saying ``false''---is trivially gamed whenever the
evaluation corpus consists only of corrupted statements, since a constant refusal then scores
perfectly. The second is that reinforcement learning against a narrow skepticism objective is
destructive to the broad reasoning ability the model already has, and the damage does not announce
itself on the training objective. Both problems are problems of data and of measurement before they
are problems of algorithm.

We address the first by training on matched pairs rather than on corrupted statements alone, and by
evaluating on a balanced split in which the majority-class shortcut earns exactly chance. We address
the second by measuring general mathematical competence, on AIME 2026, alongside every
discrimination metric, and by treating any run that improves discrimination while degrading AIME as
a failed run regardless of its headline number. The training data themselves are generated with
GraphSynth \citep{cheng-etal-2026-graphsynth}, whose combination of a probabilistic factor graph over
attributes with decode-time hard masks and a span-synchronized verifier gives us the property we
need most: corruptions that are structurally well-formed and semantically non-trivial, so that the
model cannot separate the classes on surface cues.

This report describes the resulting model, which we call Euston
\citep{checkpoint_euston}, and states plainly both what it achieves and what remains unresolved. Our
contributions are a data-generation pipeline for paired mathematical claim verification built on
GraphSynth; a GRPO recipe with a rule-based, zero-API reward that improves discrimination by
$27.5$\pp{} without a statistically significant loss on AIME 2026 under official semantics; a
decomposition showing that roughly half of the apparent accuracy gain is attributable to the model
learning to emit a parseable verdict at all rather than to improved judgement; and evidence, from baselines that
vary the generated data at fixed optimiser and the adaptation method at fixed data---supervised
fine-tuning with and without synthetic data, and an Evol-Instruct comparison---that the
quality of the generated negatives rather than the optimiser is what decides whether discrimination
is bought at the cost of general capability.

%%%%%%%%%%%%%%%%%%%%%%%%%%%%%%%%%%%%%%%%%%%%%%%%%%%%%%%%%%%%%%%%%%%%%%%%%%%%%%%
\section{Data Generation with GraphSynth}
\label{sec:data}

The quality of a claim-verification dataset is determined almost entirely by the quality of its
negatives. Corruptions produced by unconstrained prompting of a language model tend to fail in one
of two directions: they are either so crude that a surface heuristic separates them from the
positives, or they are so aggressive that the resulting statement is malformed rather than false.
GraphSynth was designed for precisely this trade-off between diversity and reliability
\citep{cheng-etal-2026-graphsynth}. It models the space of admissible statements as a probabilistic
factor graph over attributes, samples attribute combinations from that graph to obtain coverage, and
then enforces correctness during decoding by two independent mechanisms: a schema mapping compiled
into hard masks that constrain the token distribution to syntactically valid continuations, and a
span-synchronized verifier that rejects logical contradictions as spans are emitted rather than
after the fact. The reported behaviour of the generator---near-perfect structural integrity and
approximately 94\% attribute coverage on the factor-graph solution---is what allows us to treat the
negatives as hard rather than merely wrong.

We applied this pipeline to mathematical statements extracted from arXiv papers, generating for each
source statement a matched corrupted counterpart under the same attribute assignment, so that the
two members of a pair differ in truth value and as little else as possible. The procedure yielded
3{,}026 true/false pairs, or 6{,}052 statements in total, drawn from papers with arXiv identifiers
spanning 1001 through 2512---that is, January 2010 through December 2025, with no 2026 material.

Because the evaluation sets are drawn from papers posted between February and May 2026 (identifiers
2602 through 2605), the training and evaluation corpora are temporally disjoint by construction. We
verified this disjointness explicitly along three axes rather than relying on the date filter alone:
arXiv identifier, paper title, and statement text. No overlap was found on any axis. This matters
more than the usual deduplication hygiene, because the base model's pretraining corpus plausibly
contains the source papers for the training split, and we wanted the evaluation to measure
verification rather than recall.

%%%%%%%%%%%%%%%%%%%%%%%%%%%%%%%%%%%%%%%%%%%%%%%%%%%%%%%%%%%%%%%%%%%%%%%%%%%%%%%
\section{Training}
\label{sec:training}

We fine-tuned \path{deepseek-ai/DeepSeek-R1-0528-Qwen3-8B} \citep{deepseekr1,qwen3} using Group
Relative Policy Optimization \citep{shao2024deepseekmath} as implemented in \texttt{verl}
\citep{sheng2024hybridflow}. GRPO is the natural choice here because it dispenses with the learned
value network of PPO \citep{schulman2017ppo} and normalises advantages within a group of rollouts
sampled for the same prompt, which suits a task whose reward is binary and whose difficulty varies
sharply from prompt to prompt.

The reward function is deliberately austere. For each rollout we extract the final
\verb|\boxed{...}| expression and award 1.0 if it casefold-matches the ground-truth label, True or
False, and 0.0 otherwise. No model is called at any point during training. This zero-API design
costs us the nuance that a judge model could provide, but it buys two things that matter more: the
reward cannot be hacked by writing text that flatters a judge, and the training run is exactly
reproducible from the data and the seed. Table~\ref{tab:hparams} lists the configuration in full.

\begin{table}[htbp]
\centering
\caption{Training configuration. The run consumed roughly 13.8 hours of wall-clock time.}
\label{tab:hparams}
\small
\begin{tabular}{@{}ll@{}}
\toprule
\textbf{Component} & \textbf{Setting} \\
\midrule
Base model            & \path{deepseek-ai/DeepSeek-R1-0528-Qwen3-8B} (8B, BF16) \\
Algorithm             & GRPO (\texttt{verl}) \\
KL loss coefficient   & 0.001 \\
Entropy coefficient   & 0 \\
Reward                & Rule-based, zero-API; 1.0 iff final \verb|\boxed{}| casefold-matches label \\
Learning rate         & $5\times10^{-6}$ \\
Train batch size      & 32 \\
PPO mini-batch size   & 32 \\
Rollouts per prompt   & 8 \\
Max prompt length     & 3{,}072 tokens \\
Max response length   & 16{,}384 tokens \\
Schedule              & 1 epoch $=$ 189 steps \\
Hardware              & $4\times$ H100, $\approx$13.8 h \\
Training data         & 6{,}052 statements / 3{,}026 true--false pairs (Sec.~\ref{sec:data}) \\
\bottomrule
\end{tabular}
\end{table}

Two hyperparameters deserve comment because they bear on the results in
Section~\ref{sec:results}. The KL coefficient of 0.001 is low, which grants the policy substantial
freedom to move away from the reference model; this is what makes the general-capability measurement
in Section~\ref{sec:capability} load-bearing rather than a formality. The entropy coefficient of zero
means nothing in the objective opposes the collapse of output diversity, and the rollout budget of
eight samples per prompt is the only thing supplying exploration pressure.

%%%%%%%%%%%%%%%%%%%%%%%%%%%%%%%%%%%%%%%%%%%%%%%%%%%%%%%%%%%%%%%%%%%%%%%%%%%%%%%
\section{Evaluation Protocol}
\label{sec:protocol}

We evaluate along three axes that answer three different questions, and we report all three because
any one of them in isolation is misleading.

The first axis is discrimination on a balanced held-out split of 200 true and 200 false statements,
one pair per source paper, scored by the rule-based extractor \path{scripts/confmat_broken.py}
with no model in the loop. Chance is 50\% by construction and the constant-refusal strategy scores
exactly 50\%, so the split cannot be gamed by skepticism alone. The second axis is the official
BrokenArXiv protocol, run through the MathArena harness at commit \texttt{a11194d}
\citep{matharena} using \path{scripts/run.py} and \path{scripts/judge/judge.py} with
\path{gemini-3.1-pro} as judge under the \path{configs/judges/arxiv_judge_post_march}
configuration and \path{judge_points_max}~$=2$. The third axis is capability retention, measured
on the 30 problems of AIME 2026.

All generations use temperature 0.6, top-$p$ 0.95, $n=4$ samples per problem, and a maximum of
32{,}768 tokens. Confidence intervals come from a problem-level paired bootstrap with $B=20{,}000$
resamples and seed 20260726.

Because long-form reasoning runs sometimes exhaust the token budget before producing an answer, AIME
results are reported as a pair of bounds. The lower bound follows official semantics and scores a
dropped run as wrong; the upper bound scores only the runs that landed. The two bounds coincide when
nothing is truncated and diverge in proportion to the truncation rate, which differs substantially
between the models we compare, so the choice of bound is itself a source of apparent disagreement
and we give both.

%%%%%%%%%%%%%%%%%%%%%%%%%%%%%%%%%%%%%%%%%%%%%%%%%%%%%%%%%%%%%%%%%%%%%%%%%%%%%%%
\section{Results}
\label{sec:results}

Throughout this section, ``base'' denotes the unmodified
\path{DeepSeek-R1-0528-Qwen3-8B}; ``Euston'' denotes the released model, trained on the
GraphSynth-generated pairs of Section~\ref{sec:data}; and ``Euston-pilot'' denotes an earlier run of the same
recipe on a smaller GraphSynth corpus, retained here as a baseline rather
than as a controlled ablation.

\subsection{Discrimination on the balanced split}

\begin{figure}[htbp]
\centering
\includegraphics[width=\textwidth]{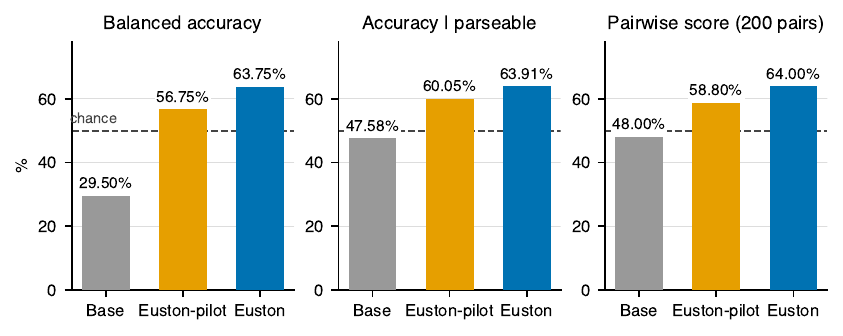}
\caption{Verdict quality on the balanced held-out split (200 true, 200 false; one pair per source
paper). Dashed lines mark chance. The left panel scores unparseable outputs as wrong, the middle
panel conditions on a parseable verdict, and the right panel reports the pairwise score over all
200 pairs with ties credited 0.5.}
\label{fig:verdict}
\end{figure}

Balanced accuracy on the held-out split rises from 29.50\% for the base model to 63.75\% for Euston,
a gain of 34.25\pp{} (Table~\ref{tab:balanced}, Figure~\ref{fig:verdict}). The base figure is below
chance, which invites misreading: the base model is not anti-correlated with the truth, it is
frequently silent. On 38.0\% of items it produces no parseable verdict within the token budget, and
those items are scored as wrong. Restricting attention to items where a verdict was actually emitted,
the base model scores 47.58\%, which is chance to within noise, and Euston scores 63.91\%.

This decomposition is the honest way to read the headline number. Of the 34.25\pp{} gain,
approximately 18\pp{} is attributable to the model learning to terminate with a verdict rather than
running out of budget, and 16.3\pp{} to improved judgement. Both components are real improvements
and both are consequences of the training, but only the second is discrimination in the sense the
benchmark intends to measure.

The discrimination gap---defined as $P(\text{say false} \mid \text{false}) - P(\text{say false} \mid
\text{true})$, and therefore immune to any constant shift in skepticism---moves from $-0.5$\pp{}
($z=-0.1$) for the base model to $+27.5$\pp{} ($z=+6.0$) for Euston. The base model, in other words,
carries essentially no information about the truth of the statement it is shown; Euston carries a
substantial and highly significant amount. The positive likelihood ratio improves correspondingly
from 0.99 to 1.53.

\begin{table}[htbp]
\centering
\caption{Balanced held-out split: 200 true and 200 false statements, one pair per source paper.
The discrimination gap is $P(\text{say false}\mid\text{false}) - P(\text{say false}\mid\text{true})$.}
\label{tab:balanced}
\small
\begin{tabular}{@{}lccc@{}}
\toprule
\textbf{Metric} & \textbf{Base} & \textbf{Euston-pilot} & \textbf{Euston} \\
\midrule
Balanced accuracy (chance $=$ 50\%) & 29.50\% & 56.75\% & \best{63.75\%} \\
\quad among parseable verdicts only & 47.58\% & 60.05\% & \best{63.91\%} \\
Discrimination gap                  & $-0.5$\pp{} & $+19.0$\pp{} & \best{$+27.5$\pp} \\
\quad significance                  & $z=-0.1$ & $z=+3.9$ & $z=+6.0$ \\
Positive likelihood ratio (LR$+$)   & 0.99 & 1.42 & \best{1.53} \\
False-rejection rate on TRUE statements & 38.5\% & 45.5\% & 51.5\% \\
No parseable verdict (truncated)    & 38.0\% & 5.5\% & \best{0.25\%} \\
\bottomrule
\end{tabular}
\end{table}

One number in Table~\ref{tab:balanced} moves in the wrong direction and should not be passed over.
The false-rejection rate on true statements climbs from 38.5\% to 51.5\%, meaning that Euston now
rejects a majority of the correct statements it is shown. Discrimination improved because the
rejection rate on false statements grew faster still, not because the model became careful. A
verifier that rejects half of all valid mathematics is not yet useful as a filter, and closing this
gap---rather than widening the discrimination gap further---is the obvious next target.

\subsection{Pairwise adjudication}

The intended deployment presents both members of a pair and asks which is the corrupted one, so we
also score pairwise (Table~\ref{tab:pairwise}). Here the model's behaviour is best described as
selective prediction. It assigns different verdicts to the two members of a pair---that is, it
commits---on 41\% of pairs, up from 21\% for the base model, and when it commits it is right 84.1\%
of the time, against 40.5\% for the base model. Crediting ties at 0.5 gives an overall pairwise
score of 64.0\% across all 200 pairs, up from 48.0\%.

\begin{table}[htbp]
\centering
\caption{Pairwise adjudication over 200 matched pairs. Coverage is the fraction of pairs receiving
different verdicts for the two members; accuracy within coverage conditions on those pairs.}
\label{tab:pairwise}
\small
\begin{tabular}{@{}lccc@{}}
\toprule
\textbf{Metric} & \textbf{Base} & \textbf{Euston-pilot} & \textbf{Euston} \\
\midrule
Coverage (pairs given different verdicts) & 21\% & 42\% & 41\% \\
Accuracy within coverage                  & 40.5\% & 71.1\% & \best{84.1\%} \\
Pairwise score, all 200 pairs (ties $=$ 0.5) & 48.0\% & 58.8\% & \best{64.0\%} \\
\bottomrule
\end{tabular}
\end{table}

The 84.1\% figure is the most favourable number in this report and also the most easily
over-claimed. It is conditional on the model's own decision to commit, and the model abstains on
roughly 59\% of pairs. Read correctly, it says that Euston has learned a usable internal signal for
when it knows, which is a genuinely desirable property for a verifier, but it does not say that the
model adjudicates 84\% of arXiv statement pairs correctly.

\subsection{Capability retention}
\label{sec:capability}

\begin{figure}[htbp]
\centering
\includegraphics[width=0.48\textwidth]{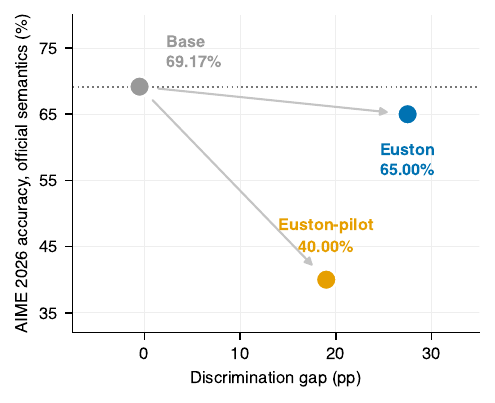}
\caption{The capability--discrimination plane. Both training runs move right, acquiring
discrimination. Euston-pilot falls off the capability axis; Euston does not. The two runs differ
in the volume of generated data they were trained on, not in the optimiser.}
\label{fig:tradeoff}
\end{figure}

The central claim of this report is not that discrimination improved, which is unsurprising under
direct optimisation, but that it improved without destroying the model. Figure~\ref{fig:tradeoff}
shows the two runs in the plane that matters.

Under official AIME semantics, Euston scores 65.00\% against the base model's 69.17\%, a difference
of $-4.17$\pp{} with $\ci{-10.83}{+2.50}$, which is not statistically significant
(Table~\ref{tab:aime}). Euston-pilot, trained with the same recipe on a smaller GraphSynth
corpus, scores 40.00\%, a loss of 29.17\pp{} that is unambiguously significant and that renders that
checkpoint unusable for anything. We regard this contrast as the most consequential observation in
the study, and we return to it in Section~\ref{sec:discussion}.

\begin{table}[htbp]
\centering
\caption{AIME 2026 (30 problems, $n=4$ samples each). The lower bound follows official semantics and
scores dropped runs as wrong; the upper bound conditions on runs that landed.}
\label{tab:aime}
\small
\begin{tabular}{@{}lccl@{}}
\toprule
\textbf{Scoring} & \textbf{Base} & \textbf{Euston} & \textbf{Difference} \\
\midrule
Lower bound (official semantics) & 69.17\% & 65.00\% & $-4.17$\pp{}, $\ci{-10.83}{+2.50}$, n.s. \\
Upper bound (landed runs only)   & 89.51\% & 71.26\% & $-12.96$\pp{}, $\ci{-21.91}{-4.94}$, sig. \\
Runs dropped                     & 25.8\% & 8.3\% & --- \\
\bottomrule
\end{tabular}
\end{table}

The two bounds disagree, and the disagreement is informative rather than a nuisance. Under the upper
bound, which asks how good the answers are given that an answer was produced, Euston is
significantly worse than the base model: 71.26\% against 89.51\%, a loss of 12.96\pp{} with
$\ci{-21.91}{-4.94}$. The reconciliation is in the last row. The base model drops 25.8\% of its AIME
runs to truncation and Euston drops only 8.3\%, so the base model's upper bound is computed on a
heavily filtered subset consisting disproportionately of problems it found easy enough to finish.
Euston attempts and completes harder problems, dilutes its own conditional accuracy, and ends up
roughly level on the metric that counts every problem. The honest summary is that Euston is
approximately as useful as the base model on AIME while being more reliable about returning
something, and that some real degradation in per-answer quality is present and is masked by the
improved completion rate.

\subsection{Generation budget}

\begin{figure}[htbp]
\centering
\includegraphics[width=0.72\textwidth]{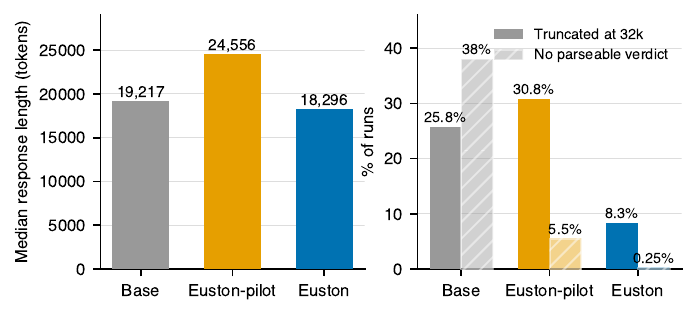}
\caption{Generation-budget behaviour. Euston improves discrimination while producing shorter
outputs than the base model and truncating far less often; Euston-pilot shows the length inflation
typical of unregularised RL on reasoning tasks.}
\label{fig:budget}
\end{figure}

The training also changed how the model spends tokens, and it changed them in the direction one
would hope for rather than the direction that usually accompanies RL on reasoning tasks
(Figure~\ref{fig:budget}). Median response length fell from 19{,}217 tokens for the base model to
18{,}296 for Euston, while Euston-pilot inflated to 24{,}556. The rate of runs truncated at the 32k limit
fell from 25.8\% to 8.3\%, and the rate of items yielding no parseable verdict on the balanced split
fell from 38.0\% to 0.25\%.

Since Euston is both shorter and better, its improvement cannot be attributed to additional
inference-time computation. Euston-pilot is longer and worse. Length inflation under reward-driven
training is a well-known pathology, and the reward used here contains no length term at all, so the
absence of inflation in Euston is fortunate rather than engineered.

\subsection{Official BrokenArXiv protocols}

On the April 2026 split (61 problems), the \texttt{sycophancy} protocol gives 19.47\% for Euston
against 14.55\% for the base model, an improvement of 4.92\pp{} with $\ci{-2.66}{+12.91}$ that is not
statistically significant at this sample size. The \texttt{disprove} protocol gives 86.89\% against
40.37\%, an apparent improvement of 46.52\pp{}.

We do not believe the second number means what it appears to mean. The official evaluation sets
contain corrupted statements only, so a model that has learned to answer ``false'' more often will
gain on \texttt{disprove} whether or not it has learned anything about mathematics. The balanced
split of Section~\ref{sec:results} exists precisely to remove this confound, and the honest estimate
of the model's discrimination is the 27.5\pp{} gap measured there, not the 46.52\pp{} measured here.
Results for the May split and the cross-period aggregate were still running at the time of writing
and are not reported.

\subsection{Data-generation and adaptation baselines}
\label{sec:ablations}

\begin{figure}[htbp]
\centering
\includegraphics[width=\textwidth]{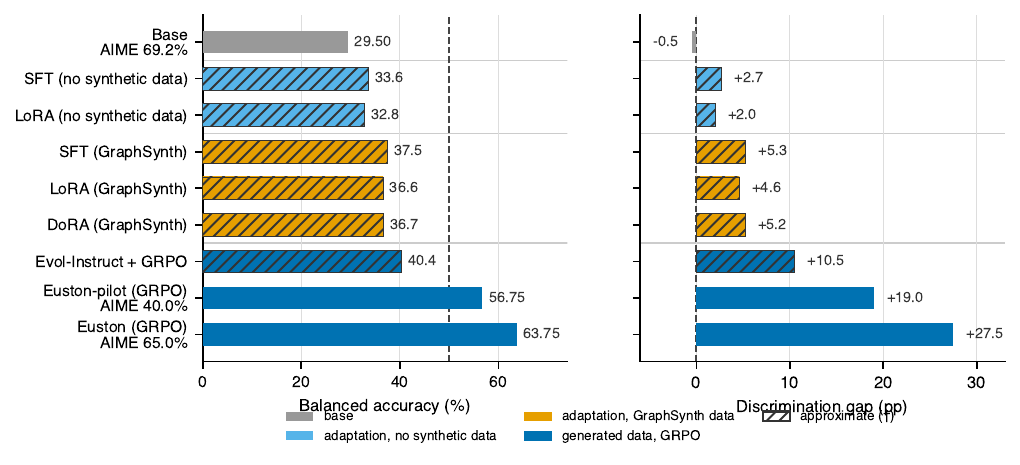}
\caption{The two ablation axes. Left, balanced accuracy on the held-out split; right, the
discrimination gap. Hatched bars are the approximate values,
reconstructed from the original run logs. Neither axis alone reaches the top of the chart: better
negatives without on-policy pressure (Panel~B) and on-policy pressure on weaker or fewer
negatives (Panel~A) both fall short of Euston.}
\label{fig:ablations}
\end{figure}

Two further comparisons bear directly on the attribution argued in
Section~\ref{sec:discussion}. Together with a no-synthetic-data control they form the design
summarised in Figure~\ref{fig:ablations}: Panel~A holds the optimiser
fixed at GRPO and varies the generated data---the generation strategy (Evol-Instruct against
GraphSynth) and, within GraphSynth, the size of the corpus---while Panel~B holds the GraphSynth
data fixed and varies the adaptation method. The full evaluation artefacts for the recalled configurations are
being regenerated; the hatched values are approximate, reconstructed from the original run
logs, and should be read as directions of effect rather than as measured effect sizes.

\paragraph{Generated data at fixed optimiser.}
In place of the factor-graph generator we built a training set of the same size by Evol-Instruct
\citep{xu2023wizardlm}, evolving seed statements through successive rounds of prompted rewriting.
Evol-Instruct raises instruction complexity, but nothing in the procedure constrains the corrupted
variant to remain structurally well-formed or to differ from its source in truth value alone, and
the resulting negatives are correspondingly easier to separate on surface cues. Training on them
improves balanced accuracy by only about ten points over the base model---below both GraphSynth
runs. The three GRPO runs therefore order Evol-Instruct, then Euston-pilot, then Euston, with the
last step reflecting the larger GraphSynth corpus rather than a different generator.

\paragraph{Adaptation method at fixed data.}
The mirror comparison is more emphatic. Supervised fine-tuning and LoRA on the base model's own
data, with no synthetic pairs at all, move discrimination by at most a couple of points: instruction
tuning teaches the format of an answer but carries no signal about truth. Supplying the GraphSynth
pairs to full SFT and to its parameter-efficient variants LoRA \citep{hu2021lora} and DoRA
\citep{liu2024dora} lifts the gain to roughly five points---so the data do help---but every
supervised variant remains far below the GRPO run, which reaches $+27.5$\pp. The group-relative
advantage is what converts matched pairs into a decision boundary; a supervised loss on verdict
labels is not a substitute for it.

Taken together the two panels bracket the recipe from both sides. Holding the optimiser fixed,
generation quality moves the result across the full range; holding the data fixed, no adaptation
method we tried substitutes for the on-policy objective. Neither factor alone reproduces Euston.

%%%%%%%%%%%%%%%%%%%%%%%%%%%%%%%%%%%%%%%%%%%%%%%%%%%%%%%%%%%%%%%%%%%%%%%%%%%%%%%
\section{Discussion}
\label{sec:discussion}

The contrast between the two runs is the clearest evidence in this study that the active
ingredient is the generated data rather than the optimiser. Euston-pilot, trained by the same recipe on a
smaller GraphSynth corpus, does acquire discrimination---its gap reaches
$+19.0$\pp{}---but it pays 29.17\pp{} of AIME accuracy for it. Euston acquires more discrimination,
$+27.5$\pp{}, and pays essentially nothing. Both runs improve on the reward the objective actually
measures, so nothing in the training-time metrics distinguishes them; the difference appears only
when general capability is evaluated separately.

There is a plausible mechanism. The reward is binary and says nothing about how a verdict should be
reached, so the policy will move wherever the data permits. Crude negatives can be separated on
surface cues, or simply by becoming more skeptical, and a policy that takes the skepticism shortcut
degrades its general mathematics as a side effect. Negatives that are structurally well-formed and
semantically non-trivial---which is what the masking and span-synchronized verification of
Section~\ref{sec:data} are for---close that shortcut off and leave the model no way to earn reward
except by actually adjudicating the statement. The baselines of Section~\ref{sec:ablations} sit where this account predicts: the
Evol-Instruct negatives are the least constrained of the three generation strategies and
discriminate the least, while supervised fine-tuning on the GraphSynth pairs---with or without a
low-rank adaptor---never approaches the on-policy run. On this
reading, the generated data---its quality and its volume---is not a detail of dataset construction
but the thing that determines whether discrimination is bought out of capability or acquired
alongside it. The practical implication is unchanged either way: a
general-capability evaluation must run periodically during training rather than once at the end, so
that divergence is caught rather than discovered after 13.8 GPU-hours.

A second observation concerns what the model actually learned. The decomposition of the accuracy
gain into roughly 18\pp{} of parseability and 16.3\pp{} of judgement suggests that a substantial part
of what GRPO taught this model is a format discipline: terminate, commit, and put the answer in the
box. That is a real capability with real value, given that the base model failed to answer 38\% of
items, but it is not the capability the benchmark name suggests, and a reader who sees only the
29.50\%-to-63.75\% headline will overestimate the epistemic component by about a factor of two. The
same caution applies in the other direction to the 84.1\% within-coverage pairwise accuracy, which
reflects a genuine and useful ability to recognise when the model knows, on the 41\% of pairs where
it commits.

Finally, the direction of the errors is worth naming. Euston rejects 51.5\% of true statements. The
model did not become a better mathematician so much as a more skeptical one, and it discriminates
better only because its skepticism is now correlated with falsity rather than uncorrelated. For
deployment as a screening filter over a literature that is mostly correct, this is the binding
constraint, and it is not addressed by further optimisation of the current objective.

%%%%%%%%%%%%%%%%%%%%%%%%%%%%%%%%%%%%%%%%%%%%%%%%%%%%%%%%%%%%%%%%%%%%%%%%%%%%%%%
\section{Limitations}
\label{sec:limitations}

The base-rate problem bounds the practical value of this checkpoint more tightly than any accuracy
figure suggests. On a realistic corpus in which roughly 1\% of statements are wrong, only about 1.5\%
of the model's ``this is false'' flags would be correct. The benchmark's balanced design is
appropriate for measuring discrimination and inappropriate as a model of deployment, and a
$+27.5$\pp{} gap is simply not enough separation to overcome a 99:1 prior.

The all-false composition of the official evaluation sets means the \texttt{disprove} numbers admit a
degenerate solution, as discussed above. Judge noise on the protocols mediated by
\path{gemini-3.1-pro} is estimated at 0.2 to 1.6\pp{}, so we treat no difference below 2\pp{} on
those protocols as meaningful; the 4.92\pp{} \texttt{sycophancy} improvement clears that threshold
but not the bootstrap confidence interval. The May and cross-period results are outstanding, so all
official-protocol conclusions here rest on 61 problems.

The AIME retention claim depends on the choice of scoring bound. Under official semantics the
regression is not significant; under the landed-runs-only bound it is, at $-12.96$\pp{}. We have
given our reasons for regarding the official bound as the fairer comparison, but a reader who
weights per-answer quality more heavily than completion reliability should reach a less favourable
verdict. The AIME sample is also 30 problems, which is small.

The comparison against Euston-pilot is not a controlled ablation. The two runs differ in the size
of the GraphSynth corpus they were trained on, and with one run per corpus size we cannot separate
that effect from ordinary run-to-run variance; the attribution in Section~\ref{sec:discussion}
rests on the mechanism we describe rather than on a controlled experiment. A clean decomposition
would vary the seed at fixed data and vary the corpus size at fixed seed, and we ran neither. The baselines of
Section~\ref{sec:ablations} address the second axis in design but not yet in evidence: their
evaluation artefacts were lost, so the values we report there are recalled to within a few points
and marked as approximate. They carry a direction but not an effect size until those runs are
repeated. Reporting Euston
rather than Euston-pilot is additionally a selection made after seeing the evaluation results, and
the reader should discount accordingly.

%%%%%%%%%%%%%%%%%%%%%%%%%%%%%%%%%%%%%%%%%%%%%%%%%%%%%%%%%%%%%%%%%%%%%%%%%%%%%%%
\section{Conclusion}

Training an 8B reasoning model to doubt false mathematics without also breaking its ability to do
mathematics is achievable, and Euston does it: balanced accuracy of 63.75\%
against 29.50\% for the base model, a discrimination gap of $+27.5$\pp{} at $z=+6.0$, AIME 2026
retention within noise under official semantics, and shorter outputs with a quarter of the
truncation rate. The data pipeline built on GraphSynth supplies the matched pairs that make this
measurable, and the rule-based zero-API reward makes the training exactly reproducible. What we do
not have is a controlled decomposition of why it works: the earlier pilot run, which lost
29.17\pp{} on AIME, differed in the size of its generated corpus as well as in its outcome, so the evidence points
at generation quality as the active ingredient without isolating it. Establishing that, alongside
periodic capability evaluation, general-mathematics replay, and a tighter KL constraint, is the
natural next experiment.

%%%%%%%%%%%%%%%%%%%%%%%%%%%%%%%%%%%%%%%%%%%%%%%%%%%%%%%%%%%%%%%%%%%%%%%%%%%%%%%
\section*{Reproducibility}
\addcontentsline{toc}{section}{Reproducibility}
\sloppy

Euston is an 8B-parameter checkpoint in BF16, released under the MIT licence inherited from the
base model. Training used \path{MathArena/brokenarxiv-training}; evaluation used
\path{MathArena/brokenarxiv-0426} and \path{MathArena/brokenarxiv-0526}. Official-protocol
evaluation used the MathArena harness at commit \texttt{a11194d}
(\path{scripts/run.py}, \path{scripts/judge/judge.py}) with judge \path{gemini-3.1-pro},
configuration \path{configs/judges/arxiv_judge_post_march}, and \path{judge_points_max}~$=2$.
The balanced split was scored with \path{scripts/confmat_broken.py}, which calls no model.
Sampling used temperature 0.6, top-$p$ 0.95, $n=4$, and \path{max_tokens} 32{,}768. Confidence
intervals used a problem-level paired bootstrap with $B=20{,}000$ and seed 20260726. Euston-pilot is retained for comparison only and is not recommended for use.

\vspace{1em}
\bibliography{references}

\end{document}